# Predicting Heart Failure Readmission from Clinical Notes Using Deep Learning

Xiong Liu[1], Yu Chen[2], Jay Bae[2], Hu Li[2], Joseph Johnston[2], Todd Sanger[1]
[1]Advanced Analytics and Data Sciences, Eli Lilly and Company, Indianapolis, Indiana, USA
[2]Lilly Research Laboratories, Eli Lilly and Company, Indianapolis, Indiana, USA
Email: liu_xiong@lilly.com

*Abstract*—Heart failure hospitalization is a severe burden on healthcare. How to predict and therefore prevent readmission has been a significant challenge in outcomes research. To address this, we propose a deep learning approach to predict readmission from clinical notes. Unlike conventional methods that use structured data for prediction, we leverage the unstructured clinical notes to train deep learning models based on convolutional neural networks (CNN). We then use the trained models to classify and predict potentially high-risk admissions/patients. For evaluation, we trained CNNs using the discharge summary notes in the MIMIC III database. We also trained regular machine learning models based on random forest using the same datasets. The result shows that deep learning models outperform the regular models in prediction tasks. CNN method achieves a F1 score of 0.756 in general readmission prediction and 0.733 in 30-day readmission prediction, while random forest only achieves a F1 score of 0.674 and 0.656 respectively. We also propose a chi-square test based method to interpret key features associated with deep learning predicted readmissions. It reveals clinical insights about readmission embedded in the clinical notes. Collectively, our method can make the human evaluation process more efficient and potentially facilitate the reduction of readmission rates.

## I. INTRODUCTION

Heart failure is a leading cause of readmissions among the elderly populations [1]. Frequent heart failure hospitalizations pose significant burden on patients and healthcare resources [2]. The prevalence of readmission and financial penalties linked to readmission rates have intensified efforts to reduce rehospitalization [3]. Identifying patients with high readmission risk can help health care providers direct resources and services to those patients to prevent avoidable readmissions, and early prediction of the hospital readmission risk is an important step in achieving this goal [4].

Large scale adoption of Electronic Health Records (EHR) has resulted in the rapid growth in volume and diversity of medical data [5]. This has created the opportunity to enable predictive modeling to proactively identify potential hospital readmissions and improve care for patients. Several machine learning methods have been proposed to predict the risk of hospital readmission, including random forest, boosting, and support vector machines [6]. More recently, deep learning methods such as recurrent neural networks (RNN) and long short-term memory (LSTM) neural networks have been applied to enhance the performance of prediction tasks [7],[8]. However, most approaches rely on a large number of clinical variables, thereby, requiring intensive feature engineering. The most valuable and relevant information about medical conditions and hospitalization may exist only in clinical notes or narratives. Recent advances in natural language processing (NLP) and deep learning have enabled machines to learn a rich representation of medical language for efficient and accurate prediction [9]. Currently, predicting readmission using clinical notes is not well exploited in the literature. We aim to leverage NLP and deep learning to provide new perspectives on readmission.

We propose an NLP deep learning framework to predict readmission based on convolutional neural networks (CNN) [10]. Our method leverages word embeddings to represent words in the clinical notes without any feature engineering. It then uses CNN architectures to automatically generate feature maps and train models to predict whether an admission is likely to be followed by a readmission. We conduct experiments involving prediction of two outcomes: general readmission and 30-day readmission. When compared with random forest, a conventional machine learning method, the CNN approach outperforms random forest in both prediction tasks. We additionally propose a chi-square test based method to allow for the interpretation of deep learning predicted results. The results indicate that CNNs are a valuable alternative to existing methods in readmission prediction and are worthy of further investigation. Moreover, the chi-square based interpretation could be used to assist physicians during patient risk assessment.

In summary, our major contributions are as follows:
- An NLP deep learning approach to predict heart failure hospital readmission from clinical notes without tedious feature engineering.
- Experiments on both general readmission and 30-day readmission prediction demonstrate that the deep learning method achieves better performance than conventional random forest methods.
- A chi-square based method to select and interpret key features from deep learning results.
- Interpretation of deep learning results identifies medical insights to heart failure readmission.

The rest of the paper is organized as follows. We first describe materials and methods for readmission prediction. Then we present results to demonstrate the effectiveness of our prediction framework. Finally, we conclude the paper and outline future research.

## II. MATERIALS AND METHODS

MIMIC III is a freely available database of patients admitted to the intensive care unit (ICU) [11]. It covers over 58,000 hospital admissions of more than 46,000 patients spanning an 11-year period from 2001-2012. We leverage the detailed clinical and billing data, including discharge summary notes and ICD-9 codes assigned at discharge, from heart failure hospital admissions captured in MIMIC III to model CNNs for readmission prediction.

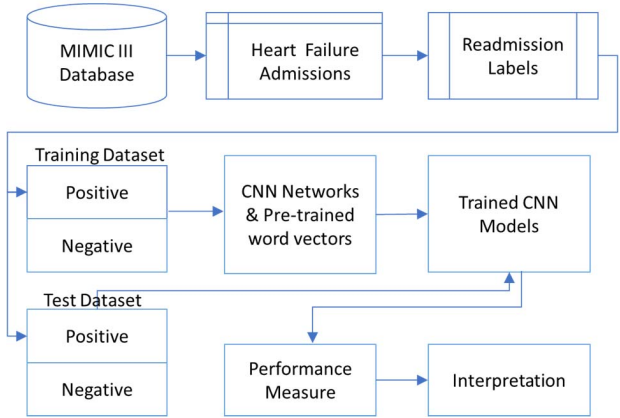

Fig. 1. Overall analysis workflow.

The overall workflow to predict readmission is shown in Fig.1. The major components include 1) retrieval of heart failure admissions from MIMIC III by ICD-9 code, 2) labeling admissions followed by any readmissions or 30-day readmissions, 3) generating a training set and a test set with balanced positive and negative samples, 4) building CNN models on the training data, 5) evaluating the models using test data, and 6) interpreting the CNN prediction results using chi-square feature analysis.

### A. Identifying Heart Failure Admissions

We use the same qualifying ICD-9 codes to identify congestive heart failure admissions as in [12],[13]. Specifically, the ICD-9 codes include 398.91, 402.01, 402.11, 402.91, 404.01, 404.03, 404.11, 404.13, 404.91, 404.93, 428.0, 428.1, 428.20, 428.21, 428.22, 428.23, 428.30, 428.31, 428.32, 428.33, 428.40, 428.41, 428.42, 428.43, and 428.9. If any of these ICD-9 codes appear in the diagnosis code, we label it as a heart failure admission.

Next, we identify the clinical notes for each admission. We focus on readmission prediction using the discharge summary notes. If an admission doesn't have a discharge summary, it is excluded; if an admission has multiple summary notes, the one with the longest text is selected.

### B. Labeling Readmissions

We take each heart failure admission as a sample and label it according to its subsequent admission if any. Fig. 2 shows the two scenarios to label general readmission and 30-day readmission. For general readmission, if an admission is followed by another admission, it is labeled positive (having readmission), otherwise negative (not having readmission). For 30-day readmission, if an admission is followed by another admission within 30-days, it is labelled positive (having 30-day readmission), otherwise negative (not having 30-day readmission).

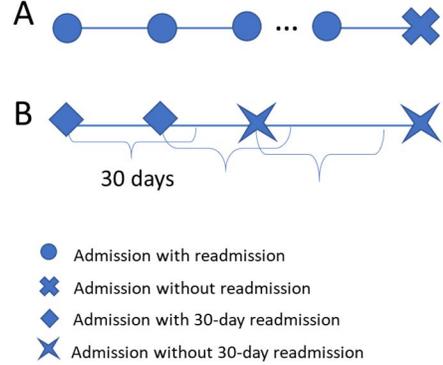

Fig. 2. Readmission labeling: A) general readmission labeling without considering the time period between admissions; B) 30-day readmission by considering the 30-day gap between admissions.

### C. Generating Training and Test sets

Given the positive readmission samples in general readmission and 30-day readmission, we randomly select the same number of negative samples for both datasets. This under sampling technique is often used to produce better predictions when the original data is unbalanced. For each dataset, we set aside 10% of the overall samples for testing of final prediction performance. We use the remaining 90% of the data to train the predictive models using an iterative 10-fold cross validation procedure.

### D. CNN Modeling

CNN is a deep learning method that uses feed-forward multi-layer neural networks [10]. CNNs were typically used for image processing [14]; they have recently achieved good results in NLP [10],[15],[16],[17]. It does not require labor intensive feature engineering by domain experts. Fig. 3 shows the CNN architecture for readmission prediction. The input to a CNN is a set of discharge clinical notes. These clinical notes are padded to a set of word sequences of the same length. Each word in a clinical note is represented as a word vector. Initializing word vectors with those obtained from unsupervised language models is common with deep learning based NLP. We use the publicly available word2vec vectors that were pre-trained on PubMed abstracts and PubMed Central full text articles [18]. Words not present in the set of pre-trained words are initialized randomly. Therefore, a clinical note is represented as a sequence of word embeddings where each word $w_i$ is projected to a $k$-dimensional embedding vector $x_i$. A text of $n$ words is represented as the concatenation of all embedding vectors $X = (x_1, x_2, \ldots, x_n)$. Misspellings, synonyms and abbreviations of an original word often have similar word embeddings, so additional curation of words is not needed.

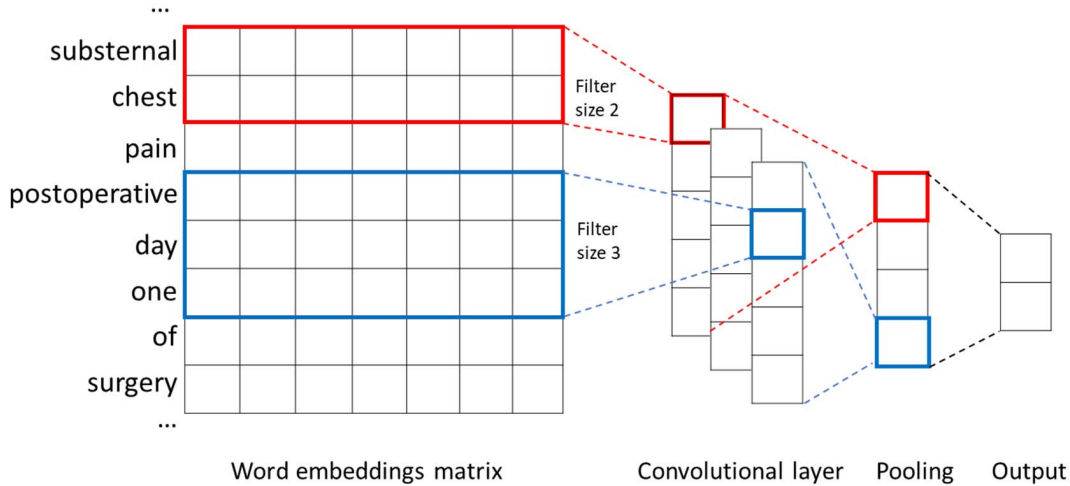

Fig. 3. CNN architecture for an example phrase. Figure adapted from Kim [10].

The embedded text is used as input to the convolutional layer where each convolutional operation applies a filter to an input window of $h$ words (filter size = $h$) to produce a new feature. This filter is applied to multiple windows of words in the word sequence to generate a feature map $C = [C_1, C_2, ..., C_{n-h+1}]$. Then a max-over time pooling operation is applied over the feature map to get the maximum value $\hat{C} = \max\{C\}$ as the feature for the filter [10]. So, the pooling scheme is able to capture the most important feature for a word sequence regardless of its length. We use multiple filters with varying window sizes of 1, 2 and 3 to obtain multiple features. These features are passed to a fully connected softmax layer which outputs the probability over labels.

Assume we have $m$ filters, the combination of these filters of varying length results in multiple outputs that form the penultimate layer $Z = [\hat{C}_1, \hat{C}_2, ..., \hat{C}_m]$. A final probability of the text label is computed as:

$$y = W \cdot Z + b,$$

where $W$ and $b$ are trainable parameters. We train a CNN for each prediction task to maximize the log-likelihood $L$ of the training data for a set of parameters $\theta$:

$$L(\theta) = \sum_{i=1}^{N} \log p(y_i | X_i; \theta),$$

where $y_i$ is the output on the $i$-th input $X_i$ and $N$ is the number of the training examples.

### E. Random Forest Modeling

To compare CNN deep learning with conventional machine learning, we use random forest to build prediction models using the same training and test data. Random forest is an ensemble learning method for classification or regression by constructing a number of decision trees and combining the predictions of the individual trees [19]. We use term frequency-inverse document frequency (TF-IDF) weights of individual words in the training set as the features into the random forest model. We tune the number of features to find the best models.

### F. Model Evaluation

We use precision, recall and F1 score to evaluate the prediction models. The training sets are used for setting parameters in CNN and the validation sets are used for selecting models. The test sets are used to obtain the final performance scores reported in the paper.

### G. Chi-square Feature Analysis

Despite the quantitative evaluation, the interpretability of deep learning results poses a challenge [16],[17]. We propose a chi-square test based feature analysis to interpret the prediction results by CNN. Originally, chi-square is a feature selection method used in machine learning:

$$\chi^2 = \sum_{k=1}^{n} \frac{(O_k - E_k)^2}{E_k},$$

where $O_k$ is the observed frequency of class, and $E_k$ is the expected frequency of class if there was no relationship between the feature and the target, and $n$ is the number of pairs of observed and expected counts. In our case n equals 4 because we are comparing the count of yes vs no over positive and negative samples for a given feature. The higher the value of the $\chi^2$ score, the more likely the feature is correlated with the class.

We repurpose chi-square feature selection to calculate the weights of words in distinguishing positive and negative samples. We apply chi-square scoring on correctly predicted samples to identify the top features (words). These features provide insights into the medical content related to readmission.

## III. EXPERIMENTS

### A. Datasets

We create two datasets: one for general readmission and the other for 30-day readmission. The statistics of the datasets are shown in Table I. As can be seen, most admissions included discharge summaries, so we did not lose many cases by concentrating on admissions with discharge summaries.

The datasets include the admissions and associated discharge summaries. Typical content in the discharge summary notes include: history of present illness, past medical history, allergies, medications, social history, physical examination, laboratory data, hospital course, discharge instructions, discharge medications, discharge condition, diagnosis, and discharge instructions. We remove the stop words and numbers and include the whole content of the discharge summary as inputs to CNNs without feature engineering.

TABLE I. STATISTICS OF HEART FAILURE ADMISSIONS

|  | # Admissions | # Admissions with discharge summaries |
|---|---|---|
| All admissions | 14,040 | 13,746 |
| Admissions followed by readmissions | 3,604 | 3,543 |
| Admissions followed by 30-day readmissions | 969 | 962 |

### B. Experimental Setup

The implementation of CNN used Keras and TensorFlow. We used the 200-dimentional word2vec vectors of [18], which were pre-trained on PubMed abstracts and PubMed Central full text articles. We used the same parameter settings for all the classifiers. We used a 24 core RHEL 7 Linux box with a Nvidia Tesla K80 GPU to train the models.

The implementation of random forest used scikit-learn. We tested multiple number of features ranging from 10,000 to 25,000 with a step of 5,000.

### C. Experimental Results

Table II shows the experimental results of different models. We can observe that the CNN model achieves the best performance in both readmission prediction tasks, giving an F1 score of 0.756 on general readmission prediction and 0.733 on 30-day readmission prediction. In contrast, the baseline random forest model only achieves F1 scores of 0.674 and 0.656 on general readmission and 30-day readmission, respectively. This shows that the CNN approach can capture richer contextual information than the TF-IDF baseline random forest approach to distinguish positive and negative samples.

TABLE II. HEART FAILURE READMISSION PREDICTION PERFORMANCE

| Task | Model | Prec | Rec | F1 | Acc |
|---|---|---|---|---|---|
| General readmission | Deep learning (CNN) | 0.759 | 0.754 | 0.756 | 75.70% |
|  | Random forest (TF-IDF) | 0.720 | 0.633 | 0.674 | 69.35% |
| 30day readmission | Deep learning (CNN) | 0.698 | 0.771 | 0.733 | 71.88% |
|  | Random forest (TF-IDF) | 0.690 | 0.625 | 0.656 | 67.19% |

### D. Chi-squared based Feature Selection

We applied chi-square feature scoring to the correctly predicted results by CNN. Fig. 4A shows the top 20 features for general readmission by chi-square scoring. Fig. 4B shows the top 20 features for the 30-day readmission prediction. Table III shows the full names of the abbreviated terms.

TABLE III. FULL NAMES OF ABBREVIATIONS

| Abbreviation | Full name | Meaning |
|---|---|---|
| PO | Per os | medication to be taken orally. |
| Sig | signetur, or "let it be labeled" | a standard part of a written prescription that specifies directions for use of the medicine |
| HD | Hospital day |  |
| Mg | milligram | a unit of measurement of mass commonly used to designate medication dose |
| BiPAP | bilevel positive airway pressure | A non-invasive form of mechanical ventilation used in the treatment of congestive heart failure and other respiratory disorders |
| esrd | end-stage renal disease | heart failure is highly prevalent in patients with end-stage renal disease |
| mL | milliliter | a unit of measurement of volume commonly used to designate medication dose |

### E. Feature Interpretation

We further count the frequency of the key features derived by chi-square. Table IV shows the frequency of the top features in correctly predicted samples of readmission (Fig. 4A). Positive readmissions have more mentions of drug related information including tablet, po (medication to be taken orally), sig (directions for use of the medicine), mg, and torsemide (a diuretic medication used to treat fluid retention caused by congestive heart failure).

Positive readmissions also have more mentions of BiPAP (bilevel positive airway pressure) which is used to treat heart failure [20], and ESRD (end-stage renal disease). There are also more mentions of hd (hospital days), campus, parking, garage, showing more activities during hospital course.

Non-readmissions have more mentions of operations related terms such as postoperative and (cardiac) catheterization. "postoperative" is often mentioned in the hospital course section. It often implies the patient had a surgery on a certain day. Cardiac catheterization is a procedure used to diagnose and treat certain cardiovascular conditions. Non-readmission also has more mentions of electrocardiogram and coronary.

Table V shows the term frequency for key features associated with 30-day readmission (Fig. 4B). Positive 30-day readmissions have more mentions of drug related information including ml (drug dosage), chewable, sig (directions for use of the medicine), solution, torsemide (a pill that It can treat fluid retention caused by congestive heart failure), carvedilol (a drug used to treat high blood pressure and heart failure), and prednisone (used as an anti-inflammatory or an immunosuppressant medication).

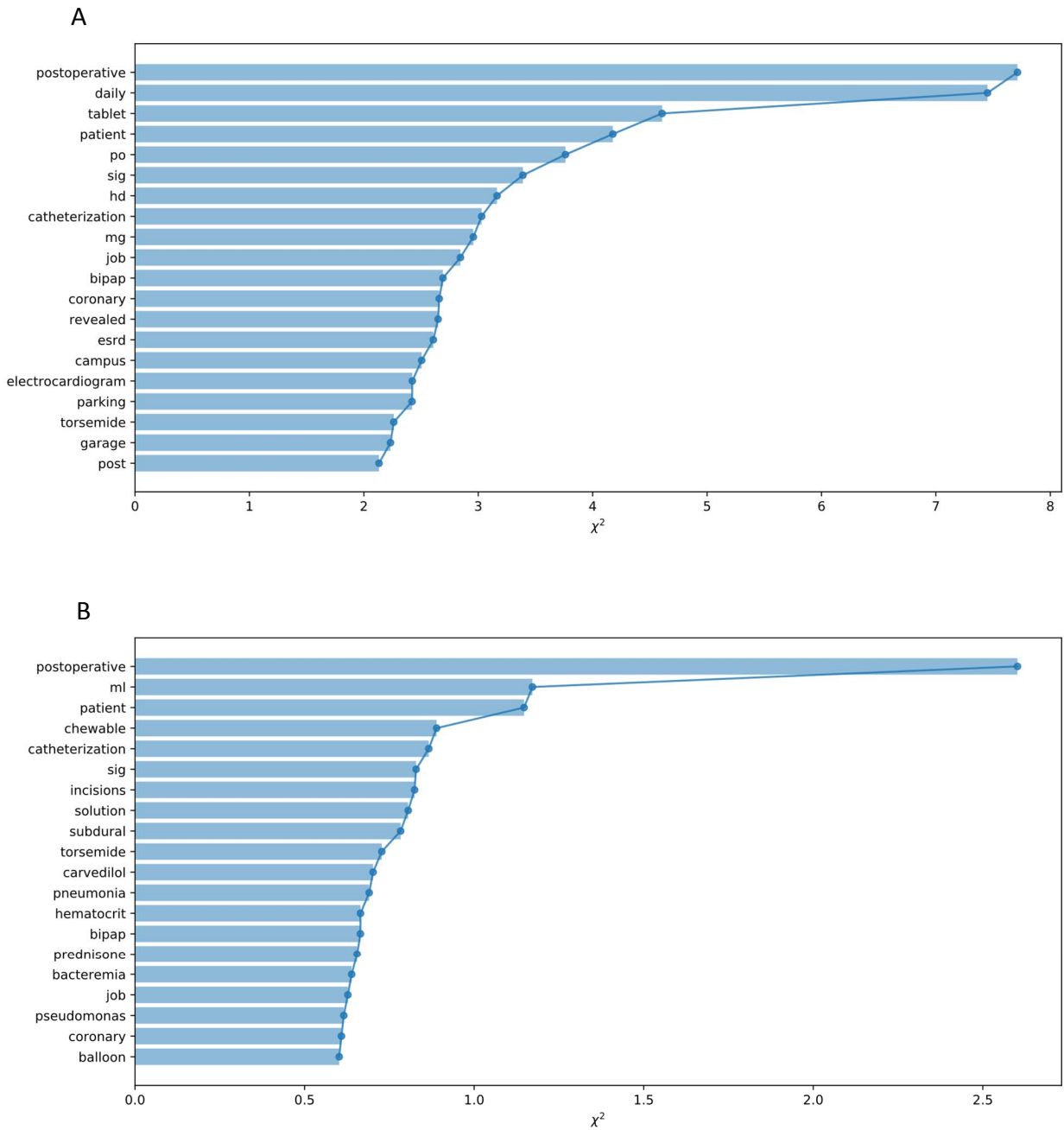

Fig. 4. Key features in correctly predicted samples for A) genernal readmission, and B) 30-day readmission.

Positive 30-day readmissions also have more mentions of heart failure related complications such as pneumonia, bacteremia, and pseudomonas. Other more frequent mentions in positive samples include BiPAP, which was shown effective in treating heart failure [20].

Non-30 day readmissions have more mentions of operations related terms such as postoperative, incisions, (cardiac) catheterization, and (intra-aortic) balloon (pump). Non-30 day readmissions also have more mentions of hematocrit, which is a blood test that measures the volume percentage of red blood cells in blood, coronary (artery disease), or coronary related exam during hospital discourse, e.g., "revealed left main coronary artery with ulcerated 90% thrombotic stenosis".

Overall, these findings suggest that clinical terms associated with greater disease severity or intensity of care are associated with increased likelihood of readmission, whereas terms indicating procedures for potential reversable conditions (e.g., cardiac catheterization) are associated with a decrease of likelihood of readmission.

TABLE IV. TOP FEATURES IN CORRECTLY PREDICTED SAMPLES OF GENERAL READMISSION

| Term | Count in positive samples (n=267) | Count in negative samples (n=269) |
|---|---|---|
| postoperative | 27 | 366 |
| daily | 3941 | 1326 |
| tablet | 5580 | 2620 |
| patient | 3028 | 4336 |
| po | 4200 | 1922 |
| sig | 3400 | 1582 |
| hd | 335 | 53 |
| catheterization | 112 | 368 |
| mg | 7042 | 3639 |
| job | 48 | 252 |
| bipap | 199 | 19 |
| coronary | 315 | 663 |
| revealed | 140 | 418 |
| esrd | 140 | non-top2000 |
| campus | 135 | non-top2000 |
| electrocardiogram | non-top2000 | 90 |
| parking | 135 | non-top2000 |
| torsemide | 97 | non-top2000 |
| garage | 121 | non-top2000 |
| post | 310 | 640 |

TABLE V. TOP FEATURES IN CORRECTLY PREDICTED SAMPLES OF 30-DAY READMISSION

| Term | Count in positive samples (n=74) | Count in negative samples (n=64) |
|---|---|---|
| postoperative | non-top2000 | 79 |
| ml | 339 | 76 |
| patient | 767 | 952 |
| chewable | 67 | non-top2000 |
| catheterization | 25 | 89 |
| sig | 958 | 407 |
| incisions | non-top2000 | 36 |
| solution | 125 | 16 |
| subdural | non-top2000 | 25 |
| torsemide | 45 | non-top2000 |
| carvedilol | 39 | non-top2000 |
| pneumonia | 161 | 34 |
| hematocrit | 35 | 85 |
| bipap | 41 | non-top2000 |
| prednisone | 87 | 12 |
| bacteremia | 40 | non-top2000 |
| job | 12 | 44 |
| pseudomonas | 32 | non-top2000 |
| coronary | 74 | 141 |
| balloon | non-top2000 | 28 |

IV. CONCLUSIONS

We have proposed an NLP deep learning approach to automatically predict general readmission and 30-day readmission for heart failure on clinical notes. Unlike conventional methods that rely on structured data and feature engineering, our method is novel in using clinical notes and deep learning without feature engineering. The result is promising with higher prediction performance than conventional random forest models. We also developed a chi-square based feature analysis to automatically retrieve key features from clinical notes. These features provide clinical insights into the patterns in readmission vs non-readmission cases. Our methods of constructing a CNN network and identifying key features from clinical notes could be easily applicable to other areas with a suitable dataset. Future work may include 1) comparative study with other reported methods, potentially including both structured and unstructured data, in readmission prediction; 2) testing different word embeddings trained from clinical notes vs PubMed; 3) testing the effect of different features on random forest; 4) more methods to interpret the deep learning results; and 5) other NLP applications using CNNs.

ACKNOWLEDGMENT

We thank Sagar Kalola for assistance with database management and Jeffery Kriske for assistance with GPU computing. We also thank Adam West for helpful discussions.